# When Does an Added Planning Capability Create Operational Value? An Observable Evidence Standard

**Yihua Xu**
Independent Researcher
San Jose, California, USA
yxu442@gmail.com

**Yan Lyu and Jize Li**
Independent Researchers
Boston, Massachusetts, USA
lyu.yan@northeastern.edu; jizel@bu.edu

**Jiani He**
Independent Researcher
Seattle, Washington, USA
jianihe@alum.mit.edu

**You Lin and Yidan Hu**
Independent Researchers
New York, New York, USA
yl2356@cornell.edu; lenahu823@gmail.com

**Rubing Qiang**
Independent Researcher
Austin, Texas, USA
rubing.qiang@gmail.com

## Abstract

Industrial artificial intelligence (AI) planning systems combine predictive models, optimization logic, controls, and governance rules, but improvement in the complete policy does not show whether one addition made an independent contribution or whether that contribution justifies its burden. We address this attribution problem with an observable evidence standard that separates opportunity to act, executed-plan change, operational consequence, equivalence, uncertainty, and operating scope. Using forecast-to-plan replay on Favorita, M5, and Walmart data, we apply the standard to heterogeneous planning additions.

Three limits of single-indicator evaluation emerge. First, a composite policy can conceal negligible component contribution: one control changed only 15 of 681 executed plans. Second, changing execution does not guarantee stable benefit: the evaluated multi-period capability helped in several long-lead settings but was 7.6% more costly under tight capacity. Third, indirect evidence can prioritize review but cannot complete attribution. Independent runs produced different paths in 19 of 599 cases that a rule-based check predicted would be unchanged. Although a lower-cost diagnostic matched the full-evaluation winner in only 55.0% of cases, it selected an option within the $10^{-6}$ cost-equivalence tolerance in 79.5%; the remaining cases prevented the proxy from authorizing acceptance.

These experiments establish capability-level contribution evidence; the organization completes the value decision by weighing it against implementation and sustaining burden. The resulting capability evidence record links the observations to validation, deployment scope, fallback, monitoring, redesign, and retirement decisions.

## 1 Introduction: The Challenge of Attributing Operational Value

Industrial AI is rarely deployed as a standalone predictive model. In planning applications, learned signals are converted into executable decisions through signal combinations, optimization, operational controls, and governance rules. The resulting systems are composite decision systems: an added capability may be predictive or learned, optimization-based, rule-based, control-oriented, or governance-oriented. As these layers accumulate, system-level improvement becomes harder to attribute to one addition.

In this study, an *added planning capability* is any new model, control rule, optimization logic, or governance constraint introduced into an existing planning process. Each addition creates data, computation, validation, monitoring, training, and maintenance demands. A common policy-level evaluation asks whether the complete planning policy performs well. In contrast, this study asks whether one added capability deserves operational credit and whether that credit justifies its continuing burden.

When considered in isolation, aggregate prediction scores, complete-policy costs, before–after outcomes, nominal component inclusion, and one favorable regime cannot reliably assign capability-level credit. A composite policy can improve while one component rarely acts or leaves execution unchanged. Conversely, a component can alter execution without stable benefit, and a theoretical or proxy label can misdescribe an approximate implementation. These failures motivate the central question: *How should an organization determine whether an added planning capability actually creates operational value?*

*Operational contribution* is the incremental change in execution and objective-aligned outcomes observed under the controlled capability comparison within the tested scope. *Operational credit* is the attribution of that contribution to the added capability. The organization completes the *operational-value* decision by weighing the credited contribution against implementation and sustaining burden. The intended benefit may concern cost, service, stability, risk, or governance. The public-data experiments establish controlled contribution evidence rather than enterprise return on investment or realized production savings.

### 1.1 Research Objectives and Contributions

The analysis pursues three sequential objectives. **Objective 1—Attribution:** determine whether an aggregate policy improvement belongs to the added capability itself. **Objective 2—Supported scope:** identify the operating conditions in which observed benefit meets the frozen materiality and uncertainty requirements, and where that benefit disappears or reverses. **Objective 3—Evidence sufficiency:** determine when theoretical checks or lower-cost diagnostics can reduce evaluation effort and when independent execution confirmation remains necessary. Together, these objectives convert the broad value question into an observable evidence problem.

The paper makes three connected contributions. Methodologically, it specifies an *observable capability-level evidence standard* that keeps nominal presence, opportunity to act or bind, executed-plan change, operational consequence, equivalence, uncertainty, and scope separately observable before incremental operational credit is assigned. Empirically, it demonstrates three recurring attribution failures: policy-level gains can miscredit included components, execution changes can yield regime-bounded or adverse consequences, and theoretical or proxy evidence can fail to establish implementation equivalence. Practically, the standard produces a *capability evidence record* that supports bounded lifecycle decisions.

The standard defines what must be observed, and *activation-preserving evaluation* keeps those observations visible under a frozen information boundary. Forecast-to-plan *replay* provides the controlled comparison used here. The *capability evidence record* is the structured result for one addition: it joins the frozen evaluation requirements to observed use, execution, consequence, uncertainty, and scope. The *evidence-to-action lifecycle protocol* then constrains the organizational action that record supports.

An outcome-only ablation asks whether performance changes when a component is removed. The proposed standard additionally requires evidence of whether the component had an opportunity to act, independently changed the executed path, produced a defensible or adverse consequence, remained valid within a bounded operating scope, and justified a particular lifecycle action. The methodological increment is their joint enforcement at the added-capability level under a shared execution boundary, while retaining exact ties, uncertainty, and unresolved cases across heterogeneous capability classes. The unit of analysis is the added capability rather than the complete policy.

Sections 2–4 establish the capability-level gap, evidence chain, and prespecified validation design. Section 5 answers the three objectives in sequence and translates the combined evidence into bounded lifecycle action.

## 2 Literature Review: The Capability-Level Evidence Gap

### 2.1 Prediction and Policy Performance Do Not Identify Component Value

AI-enabled demand planning now spans forecasting, decision support, execution optimization, and monitoring (Culot et al. 2024; Walter et al. 2025; Vlachos and Reddy 2025). Retail benchmarks establish performance across scale, hierarchy, and intermittency (Makridakis et al. 2022; Januschowski et al. 2022), while foundation models extend zero-shot and cross-domain forecasting (Das et al. 2024; Ansari et al. 2024). These advances sharpen the downstream question: a better predictive signal creates operational contribution only if it changes an executable decision and its consequence.

Prediction and operational outcomes can diverge. Demand-feature production planning links forecasts to service and cost (Su et al. 2025), while data-driven newsvendor work connects shifting-demand forecasts to ordering and total cost (Li et al. 2023). Predict-then-optimize and decision-focused learning connect models to constrained objectives (Elmachtoub and Grigas 2022; Mandi et al. 2024), and inventory-routing studies evaluate decisions beyond accuracy (Jia et al. 2025). These studies motivate downstream evaluation; the present task is attributing one existing element's contribution under a common execution environment.

### 2.2 Existing Foundations Leave a Capability-Level Evidence Gap

Operations research already establishes that planning value is conditional. Advanced planning separates forecasting, planning, and execution (Stadtler et al. 2015); inventory control evaluates cost and service (Silver et al. 1998; Axsäter 2015); combination can diversify forecast error (Bates and Granger 1969); and delayed flows and rolling horizons make local actions affect later states (Forrester 1961; Sethi and Sorger 1991). Value-of-modeling and algorithm-selection research also weigh added sophistication against context (Delage et al. 2014; Rice 1976).

Ablations and simulations retain useful diagnostics but do not generally require one frozen capability-level record spanning opportunity, plan change, consequence, uncertainty, scope, and lifecycle action. A cross-domain audit shows that agent-safety benchmark outcomes reflect general capability, metric formulation, or panel composition rather than the claimed property (Wang et al. 2026). This analogy motivates identification discipline; it does not validate the industrial-planning framework. The proposed standard makes the integrated requirement explicit.

Prior work shows that prediction and decision performance can diverge, planning value is conditional, and added sophistication must be judged against context (Elmachtoub and Grigas 2022; Stadtler et al. 2015; Delage et al. 2014). It usually treats the complete model or policy as the unit, offering limited guidance on whether one added capability influenced execution, where its contribution remains valid, and what lifecycle action follows.

## 3 Methods: From Policy Output to Capability Evidence

### 3.1 Capability Credit Requires a Four-Stage Evidence Chain

The observable evidence standard answers one question: what must be observed before an added capability can receive operational credit? It requires four linked observations—presence in the candidate policy, opportunity to act or bind, change to the feasible or executed plan, and a consequence relevant to the intended objective. *Activation-preserving evaluation* keeps these observations separate rather than collapsing them into a complete-policy outcome.

Here, *activation* means that the added logic is used, and *binding* means that a control restricts an otherwise available action. Equivalence is evaluated separately for internal model paths, executed plans, and operational outcomes; disagreement at one level does not establish divergence at another. Adverse evidence records harm to intended or protected outcomes, whereas unresolved evidence records insufficient exposure, counterfactual support, or precision. Thus, Presence ≠ Activation, Activation ≠ Path Change, and Path Change ≠ Benefit.

Forecast-to-plan replay instantiates the standard by holding point-in-time information, state transitions, constraints, and cost accounting fixed while the candidate capability varies. This common measurement boundary does not encode a preferred winner: inactive, tied, adverse, non-monotonic, and unresolved results remain observable. The standard defines required evidence rather than a documentation template. Figure 1 shows the evidence question before any empirical mechanism appears. It distinguishes the four empirical stages from the later organizational value decision, where observed contribution is weighed against sustaining burden.

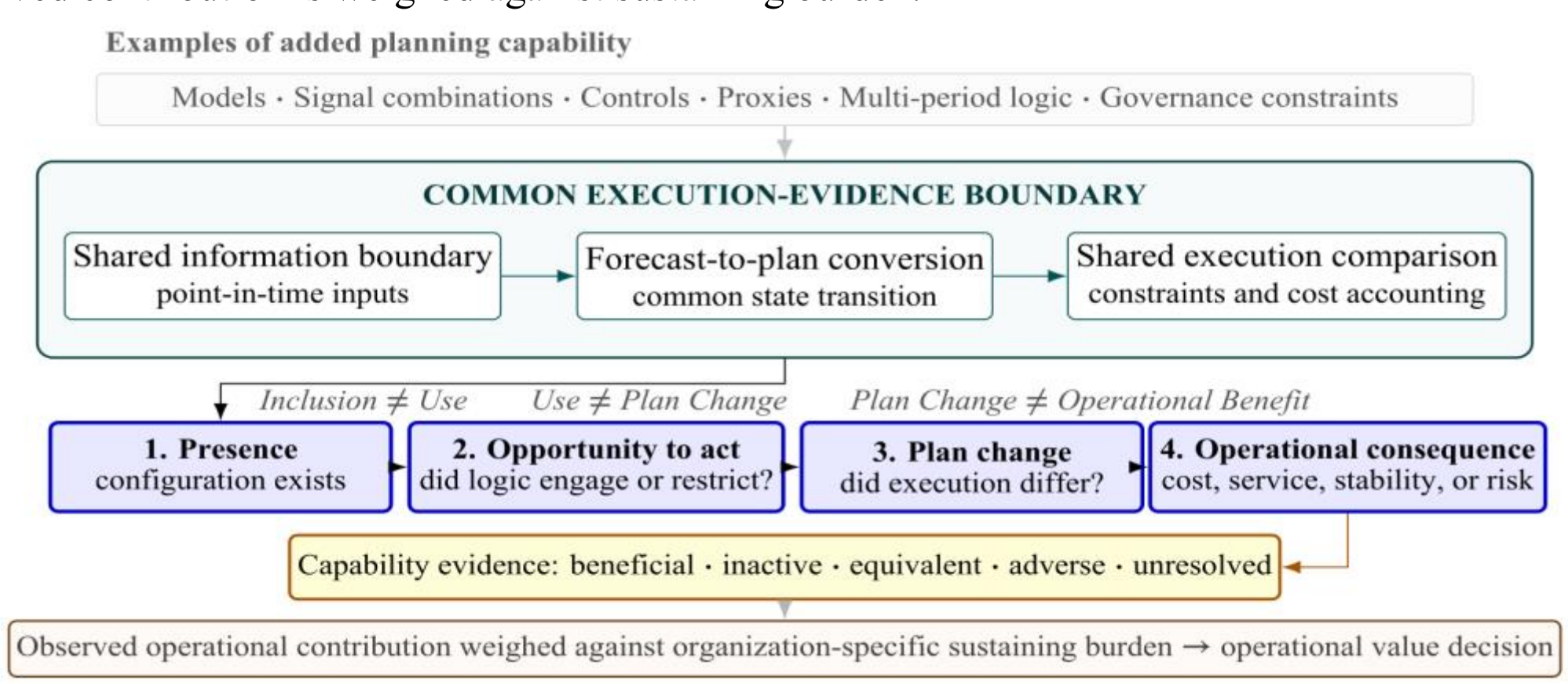

Figure 1. What evidence is required before a capability can receive operational credit? The empirical chain generates contribution evidence; the organization then weighs it against sustaining burden.

### 3.2 A Common Rolling Environment Isolates the Added Capability

All candidate capabilities are evaluated in the same rolling planning environment. In each period, a policy first selects a forecast or model using only information then available. The forecast becomes a desired order, after which capacity and optional controls determine the executed order. Realized demand is revealed only after that action is fixed, and the operational states are then updated. Equations (1)–(7) formalize this sequence. Holding it fixed attributes differences in capability use, executed plans, and outcomes to the added capability rather than to a different simulator. Let $i$, $t$, and $m$ index planning unit, period, and candidate forecast model. Forecast $\hat{y}_{i,t}^m$ is converted to order-up-to target $p_{i,t}^m$ using the forecast horizon, lead time, and validation-derived safety stock. This conversion is frozen before test.

*State and pipeline update.*

The rolling state is $x_t = (I_{t-1}, B_{t-1}, O_t, m_{t-1}, q_{t-1}, p_{t-1})$: inventory, backlog, on-order pipeline, prior model, prior order, and prior plan. For lead time $L > 0$, the first pipeline element arrives and the new order enters its end. When $L = 0$, the current order is immediately available:

$$a_t = \begin{cases} O_{t,1}, & L > 0, \\ 0, & L = 0, \end{cases} \qquad O_{t+1} = (O_{t,2}, \dots, O_{t,L}, q_t) \quad (L > 0). \tag{1}$$

Inventory position before ordering is

$$IP_t = I_{t-1} + a_t + \sum_{\ell=2}^{L} O_{t,\ell} - B_{t-1}. \tag{2}$$

*Order selection and execution.*

With safety stock $s_i$, candidate forecast $\hat{y}_t^m$ gives

$$p_t^m = (L+1)\hat{y}_t^m + s_i, \qquad q_t^* = [p_t^m - IP_t]^+. \tag{3}$$

Available capacity $C_t$ clips the desired order. To test whether an added control changes execution, relative cap $\rho$ optionally limits movement from the prior order:

$$\bar{q}_t = \min(q_t^*, C_t), \quad r_t = \rho \max(|q_{t-1}|, 1), \quad q_t = \Pi_{[\max(0, q_{t-1} - r_t),\, q_{t-1} + r_t]}(\bar{q}_t), \tag{4}$$

where $\Pi$ denotes interval projection. Without the control, $q_t = \bar{q}_t$. With $S_t = I_{t-1} + a_t + \mathbb{I}\{L = 0\} q_t$ and realized demand $y_t$ revealed only after $q_t$ is fixed,

$$I_t = [S_t - B_{t-1} - y_t]^+, \qquad B_t = [B_{t-1} + y_t - S_t]^+. \tag{5}$$

*Operational outcomes and accounting.*

Execution violation (EV) is $EV_t = [q_t^* - q_t]^+$. Inventory cost or exposure (IC) is cumulative holding plus shortage cost, and planning volatility (PV) is $\sum_t |\, q_t - q_{t-1}|$. The planning-execution gap (PEG) rate is

$$PEG(\pi) = \frac{1}{T} \sum_{t=1}^{T} \mathbb{I}\{EV_t(\pi) > 0\}. \tag{6}$$

With coefficients $h$, $b$, $c_q$, $c_e$, and $c_s$ for holding, backlog, order change, expedite, and model switching, dynamic cost is

$$C^{\text{dyn}} = \sum_t \left[ hI_t + bB_t + c_q |q_t - q_{t-1}| + c_e [q_t^* - \bar{q}_t]^+ + c_s \mathbb{I}\{m_t \neq m_{t-1}\} \right]. \tag{7}$$

The base comparative-cost weights are $h = 1.0$, $b = 5.0$, $c_q = 0.05$, $c_e = 0.10$, and $c_s = 0.05$, fixed across candidates before test; they are dimensionless, not enterprise-calibrated monetary costs. In the M5 and Walmart transfer, multi-period, and governance analyses, capacity-induced execution shortfall is reported separately rather than included in aggregate cost, so $c_e = 0$. Prespecified sensitivity analyses vary $b$ and $c_q$; governance accounting additionally varies $c_s$. The common executed path therefore supplies the observations needed after nominal presence: whether the capability could act, whether it changed execution, and whether the change affected the intended operational outcome. Every selected path is replayed under this cost. Fill rate, ending backlog, and switches remain separate outcomes so that one aggregate cost cannot hide the operational consequence.

### 3.3 Capability Cases Test Three Attribution Shortcuts

The empirical cases are not competing policies in a benchmark. Each comparison is chosen to test one shortcut by which an organization might assign operational credit too early. Table 1 connects each shortcut to the evidence required to challenge it.

Table 1. How each empirical comparison challenges a specific attribution shortcut.

| Role in the argument | Capability comparison | Mistaken inference tested | Required evidence |
|---|---|---|---|
| Attribution | Composite policy; signal combination with and without rate control | An overall policy gain belongs to every included component | Opportunity to act, executed-plan change, incremental consequence |
| Supported scope | One-step reference versus multi-period planning; frozen transfer and operating stresses | Changed execution implies stable, portable, or generally favorable value | State exposure, consequence by regime, adverse and below-threshold cases, uncertainty |
| Evidence sufficiency | Unconstrained versus budgeted planning; static diagnostic versus full execution | A theoretical status or proxy ranking establishes implementation equivalence | Independent path agreement, cost equivalence, dataset-specific P90 regret |
| Action translation | Evidence patterns across all cases | One aggregate score can support adoption | Primary status, risk overlay, supported scope, burden, unresolved findings |

The composite-policy comparison tests whether an included component inherits policy-level credit. Transfer and operating-stress comparisons then determine where an execution-changing capability remains supported. Governance and diagnostic comparisons test whether indirect evidence can substitute for execution confirmation. Read together, the resulting evidence patterns support bounded lifecycle action.

The implementations follow these purposes. Equal- and validation-weighted signal combinations isolate diversification, while wrapped and unwrapped versions isolate the rate-of-change control. A one-step dynamic greedy policy provides the reference for an approximate dynamic programming (DP) policy that carries inventory, backlog, and pipeline state across periods. A budgeted variant adds a hard switch cap $K$. The retrospective diagnostic compares one-period proxy rankings with full execution outcomes.

The lower-cost diagnostic uses normalized inventory cost (NIC), normalized volatility cost (NVC), normalized execution cost (NEC), and normalized switching cost (NSC) as one-period proxy components. With weights $\lambda_I$, $\lambda_V$, $\lambda_E$, and $\lambda_S$,

$$L^{\text{static}}(\pi) = \lambda_I NIC(\pi) + \lambda_V NVC(\pi) + \lambda_E NEC(\pi) + \lambda_S NSC(\pi). \tag{8}$$

For each M5 or Walmart unit in the strict-frozen panel, let $\hat{\pi}_i^{\text{static}}$ minimize the summed proxy over the same candidate policies used in full execution. Static-choice dynamic regret is $R_i^{\text{dyn}} = C_i^{\text{dyn}}(\hat{\pi}_i^{\text{static}}) - \min_\pi C_i^{\text{dyn}}(\pi)$, where $C_i^{\text{dyn}}$ is raw test-horizon dynamic cost. The difference is not normalized or clipped; minimization makes it nonnegative, and exact cost ties have zero regret even when winner labels differ. Choices with $R_i^{\text{dyn}} \leq 10^{-6}$ are treated as *cost-equivalent*. Choices above that tolerance are *non-equivalent*; upper-tail regret is summarized separately by dataset-specific P90 values. Because raw-cost scales differ across datasets, pooled raw percentiles are descriptive only.

Weighted absolute percentage error (WAPE) is reported separately following Hyndman and Koehler (2006). Because the rank-and-regret diagnostic uses held-out demand and execution violations, it measures retrospective triage rather than a leakage-safe production threshold. The approximate DP retains rounded states in a beam of 96; a frozen $3 \times 3$ audit checks whether the favorable long-lead and adverse tight-capacity directions persist, not tuning, convergence, or exact optimality.

### 3.4 Information Boundaries Preserve Fair Comparison

Table 2 states what each policy or diagnostic may observe. “Conversion” means operational state converts a fixed forecast into an order rather than selects the policy family. No deployable policy observes future realized demand or outcomes. The held-out information used by the static diagnostic is disclosed as retrospective only.

Table 2. Policy information access at action-selection time.

| Policy family | State access | Forecast horizon | Validation information | Realized test information | Status |
|---|---|---|---|---|---|
| Accuracy-first/fixed | Inventory, backlog, and on-order via conversion | Current | Model error | None | Deployable |
| Ensemble | Inventory, backlog, and on-order via conversion | Current | Weights only | None | Deployable |
| Rate-limit wrapper | Conversion state and prior order | Current | Cap only | None | Deployable |
| Static diagnostic | One-period proxy; no backlog or on-order state | Current | Weights only | Demand and violations | Retrospective only |
| Dynamic greedy | Inventory, backlog, and on-order | Current step | Costs and controls | None | Deployable |

| Policy family | State access | Forecast horizon | Validation information | Realized test information | Status |
|---|---|---|---|---|---|
| Approximate DP | Inventory, backlog, and on-order | Full forecast path | Costs and controls | None | Deployable |
| Budgeted DP | Same as approximate DP, plus switches | Full forecast path | Costs, controls, and $K$ | None | Deployable |
| Outcome oracle | Varies | Realized path | None | Future demand and outcomes | Oracle only |

The application programming interface (API) reveals realized test demand only after action selection, and oracle diagnostics remain separate. Candidate forecasts, transitions, and accounting are shared. Service, backlog, volatility, switching, and cost remain distinct so that agreement on one outcome does not imply overall equivalence.

Transfer evidence matters because value that appears only after local recalibration cannot be assigned to the frozen capability definition. *Validation-selected transfer* chooses a strategy using validation dynamic cost only, freezes that choice before test, and then evaluates test outcomes; the same rule applies to strict-frozen, family-recalibrated, and fully in-domain protocols, with validation ties retained. Strict frozen transfer carries equal weights and rate cap 0.40, family-frozen recalibration estimates controls from target validation, and fully in-domain tuning estimates model-family weights and controls there.

Retrospective protocol-envelope regret compares the lowest observed protocol cost with the in-domain envelope: $R^{\mathrm{env}}_{dgr} = \min_{\pi} \bar{C}_{dgr\pi} - \min_{\pi} \bar{C}_{dg,\text{in-domain},\pi}$, where $\bar{C}$ is raw test-horizon cost averaged equally over sampled units. The minimum is selected retrospectively from executed candidates. Differences are unclipped; percentages divide by the in-domain envelope, and exact ties remain zero. The statistic measures policy-family coverage rather than deployable validation-selected transfer performance.

### 3.5 Equivalence Rules Separate Inactivity from Contribution

The equivalence rules compare three levels under independent execution. A model-path difference is any disagreement in internal model choices or search paths and remains reportable, but it does not by itself earn or remove operational credit. Executed plans are equivalent only when every relevant order difference is within the $10^{-9}$ plan tolerance. Costs are equivalent when their unrounded difference is within the $10^{-9}$ cost tolerance; winner-label disagreement can therefore coexist with cost equivalence.

An implementation difference becomes operationally meaningful only when it propagates to a plan difference beyond tolerance or to an intended or protected outcome that crosses the frozen materiality or adverse-risk rule. A switching constraint is *theoretically nonrestrictive* when the independently computed unconstrained path has at most $K$ switches; expected binding means it has more than $K$. These theoretical labels never alter either branch or replace execution evidence. Any comparison lacking sufficient exposure, precision, or a required level of agreement is recorded as unresolved rather than forced into equivalence or divergence.

## 4 Experimental Design: Prespecified Tests of Attribution Failure

### 4.1 Data, Chronology, and Point-in-Time Panels

The experimental design is not a benchmark intended to identify one universally superior planning policy. It comprises prespecified tests aligned with the three objectives: component decomposition for attribution, operating-regime variation for supported scope, and independent or lower-cost comparisons for evidence sufficiency. Favorita supplies the primary daily setting, M5 adds hierarchy and intermittent demand, and Walmart adds weekly holiday and markdown stress. Formal identities, scale, panels, and roles appear once in Table 3.

Table 3. Public datasets, study scale, and distinct validation roles.

| Public dataset (short label) | Native size | Study panel | Role |
|---|---|---|---|
| Corporación Favorita Store Sales - Time Series Forecasting (Favorita) | 3,000,888 daily records; 1,782 store-family series | 100 series; 28-day test | Primary daily evaluation |
| M5 Forecasting Accuracy, the fifth Makridakis competition (M5) | 30,490 item-store series; 1,913 study-history days | 100 item-store series plus 16 department-store and 11 category-store aggregates; 28-day test | Hierarchy and intermittency transfer |
| Walmart Recruiting Store Sales Forecasting (Walmart) | 421,570 weekly records; 3,331 store-department series | 100 store-department series; 13-week test | Weekly cadence and holiday/markdown stress |

Chronological splits preserve the point-in-time boundary: Favorita and M5 use 28 validation and test days, and Walmart uses 13 validation and test weeks. The same sampled units support policy, attribution, and transfer comparisons across the selected operating conditions.

Panel identities were frozen from history available before validation and test, before model fitting or policy evaluation. Favorita retains the 100 eligible store-family series with greatest pre-validation sales after requiring at least 90 observations and 30 nonzero observations. M5 similarly retains the 100 eligible item-store series with greatest pre-validation demand after requiring at least 365 history days and 10 nonzero observations, with ties ordered by nonzero count and identifier. Walmart requires at least 80 pre-validation weeks and 20 nonzero weeks, retains the 70 highest-volume series, and samples 30 from the remaining eligible set using seed 42.

### 4.2 Frozen Evaluation Contract and Prespecified Tests

Before evaluation, the pre-evaluation contract freezes the intended objective, candidate and baseline, information and operating scope, material-benefit and adverse-risk criteria, uncertainty requirements, ownership, and reassessment rules. A numerically detectable difference is operationally meaningful only if it crosses the applicable frozen rule. In this study's unit-level cost summaries, that rule requires an effect to exceed both $10^{-6}$ and 1% of the absolute reference cost; 1% is a prespecified study threshold, not a universal deployment standard. After execution comparison, the capability evidence record combines these requirements with observed outcomes, uncertainty, unresolved findings, and organization-supplied burden.

The primary scenario fixes lead time, shortage-to-holding ratio, and moderate capacity before prespecified stresses vary carryover, capacity, switching cost, forecast horizon, and switch limits. Governance uses either $K = 2$ or a validation-derived quartile rule, both frozen before test. Uncertainty uses 1,000 paired circular time-block bootstrap replicates within planning units. Block lengths reflect each dataset's cadence, and 95% confidence intervals (CIs) accompany prespecified comparisons. Scenario labels never substitute for observed capability use, plan change, consequence, or uncertainty.

### 4.3 Validation Strategy

Each validation serves one purpose. The 28-period stress analysis identifies operating-regime effects, while the nine-setting beam/rounding audit tests whether those full-horizon directions depend on approximation settings. The tractable exact benchmark instead tests implementation fidelity by comparing exhaustive and approximate paths on 40 reduced problems. Validation-selected transfer tests frozen choices selected without test outcomes; the retrospective envelope is a secondary policy-family coverage diagnostic. Paired time-block bootstrap intervals quantify uncertainty for prespecified contrasts rather than forming a familywise hypothesis-testing screen, and the half/base/double coefficient grid tests cost-weight sensitivity. Point-in-time panel construction prevents validation or test demand from influencing unit selection.

## 5 Results and Discussion: What Conventional Evaluation Misses

### 5.1 Policy-Level Improvement Can Miscredit Components

This comparison tests whether an added component can inherit credit from a successful composite policy. Panel A of Table 4 shows lower estimated mean dynamic cost in all three datasets. The paired 95% CIs exclude zero for Favorita and M5 but include zero for Walmart. The composite effect is therefore observable, but its source is not.

Panel B decomposes that effect. Signal diversification produces the broad reduction. The rate-of-change control came into use in 16.9% of comparisons but changed only 15 of 681 executed plans, and the combined-policy effect consequently mirrors diversification. A capability embedded in a successful composite policy should receive incremental credit only when its own execution contribution is observed.

**Panel A: Composite-policy effect relative to accuracy-first**

Table 4. Why the composite-policy improvement cannot be credited to the rate-of-change control.

| **Dataset** | **Mean dynamic-cost difference** | **95% CI** |
|---|---|---|
| Favorita | −61,242 | [−81,669,−44,856] |
| M5 | −1,261 | [−2,013,−526] |
| Walmart | −35,355 | [−101,430,32,430] |

**Panel B: Component decomposition across 681 comparisons**

| **Added logic/comparison** | **Mean cost difference** | **Use rate** | **Plan-change rate** | **Attribution** |
|---|---|---|---|---|
| Signal diversification | −11,436.1 | – | 92.1% | Drives the broad composite reduction. |
| Rate-of-change control | +0.8 | 16.9% | 2.2% | Changes only 15 executed plans; no mean incremental benefit. |
| Combined policy versus accuracy-first | −11,435.2 | 16.9%[a] | 92.5% | Effect mirrors signal diversification. |

*Notes:* Panel A averages paired test-horizon cost differences equally over planning units within each primary dataset panel. Panel B's mean cost differences average 12 dataset–grain–protocol cell means equally, while its use and plan-change rates pool 681 unit comparisons; the panels therefore use different weights and are not intended to reconcile directly. [a] Use refers to the rate-of-change control within the combined policy; a plan change requires nonidentical executed orders.

## 5.2 Establishing Supported Scope Through Transfer-Coverage and Operating-Stress Evidence

Execution influence is only the beginning of attribution. Establishing a supported scope requires separate evidence of whether the capability family retains competitive coverage after transfer and whether its operational consequence remains defensible as operating conditions change.

### 5.2.1 Validation-Selected Transfer Tests Frozen Capability Definitions

The first scope test selects one strategy for each dataset, grain, and transfer protocol using validation dynamic cost only, freezes that choice, and compares its test outcome with a likewise validation-selected fully in-domain reference. Across the four dataset–grain settings, strict-frozen and family-recalibrated validation-selected policies were cost-equivalent within the study tolerance to their in-domain references, and paired 95% CIs included zero. This indicates no detected test-cost difference under the evaluated comparison; it is not proof of universal portability. Validation ties remain visible, while the retrospective envelope is retained as a secondary coverage diagnostic.

### 5.2.2 Operating Stress Bounds the Supported Scope

The second scope test asks whether a capability that changes execution remains beneficial when lead time, carryover, capacity, and switching costs change. Operational-state exposure identifies opportunity to act, not benefit. Positive pipeline carryover appeared throughout the long-lead cases, and several regimes showed favorable mean effects. However, 21% of M5 units and 47% of Walmart units remained below the predeclared material-benefit threshold.

Context also changes the direction of the observed effect. Figure 2 shows favorable but non-monotonic long-lead and switching-cost patterns, while the M5 tight-capacity case was 7.6% more costly. Across nine frozen approximation settings, the favorable long-lead and adverse tight-capacity directions persisted. Execution change alone is insufficient: the evaluated capability shows its clearest support in identified long-lead and carryover regimes and remains risk-bearing elsewhere.

The reduced benchmark takes every fifth identifier from the sorted frozen M5 panel, yielding 20 units. It uses three fixed models, eight periods, and the long-lead and tight-capacity regimes; all choices were frozen before exact outcomes were inspected. On the resulting 40 problems, 97.5% model-path agreement measures internal-path fidelity, while 100% executed-plan agreement shows that the observed path disagreement did not propagate to plan-level divergence. Median and P90 objective gaps of zero quantify outcome agreement, and the 0.081% maximum bounds the remaining discrepancy. This evidence supports fidelity only within the reduced benchmark; it does not establish full-horizon or universal equivalence.

Across the prespecified half/base/double coefficient grid, the tight-capacity reversal remained 6.5–7.9%. The diagnostic continued to show frequent cost equivalence, while non-equivalent choices retained dataset-specific upper-tail regret. Governance enforcement remained observable, although its mean cost direction varied with the comparative weights.

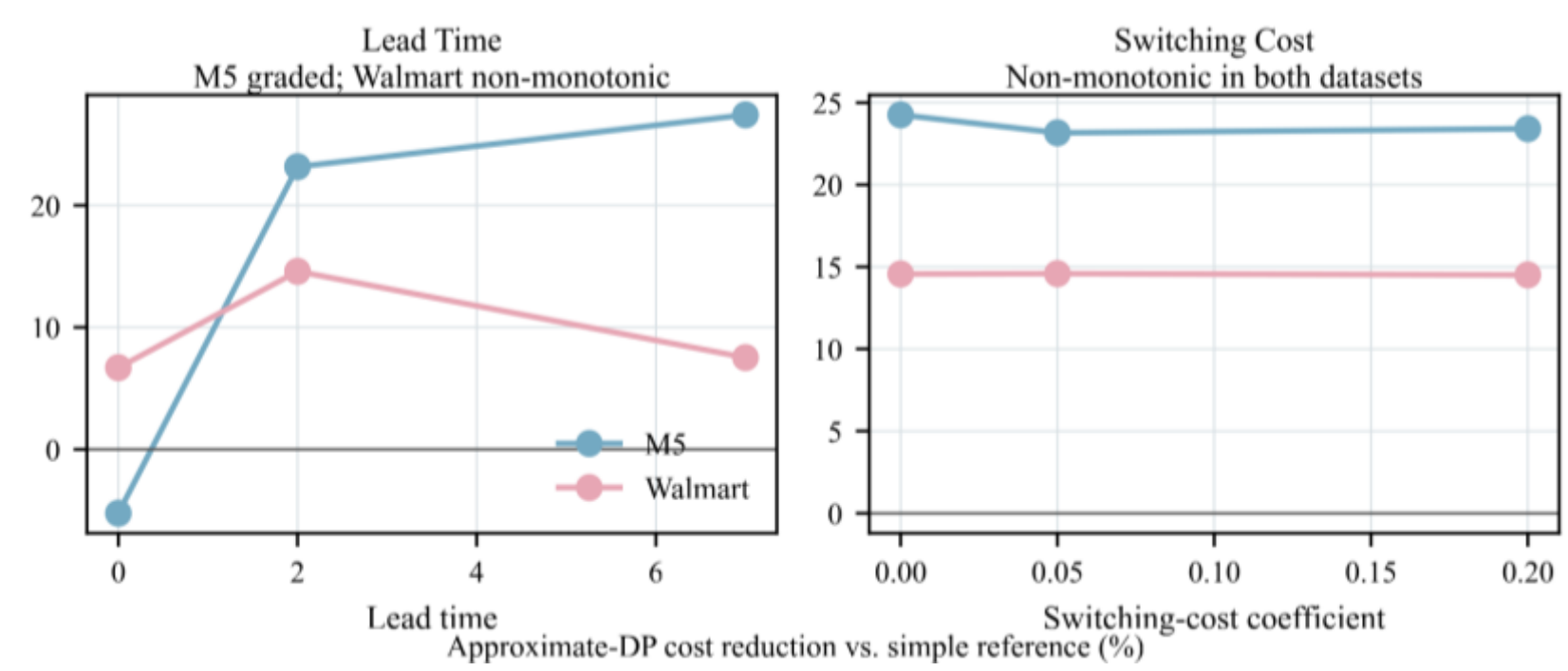

Figure 2. Multi-period value is favorable in several long-lead settings but reverses under tight capacity. Positive values indicate lower dynamic cost than the simple reference.

The stress evidence converts heterogeneous outcomes into an eligibility boundary. The evidence provides its *clearest support in identifiable long-lead and carryover regimes*, while tight-capacity and below-threshold conditions remain outside broad eligibility and retain the baseline. Once that boundary is defined, the remaining question is whether less expensive evidence can safely defer full execution confirmation.

### 5.3 Indirect Evidence Can Triage but Cannot Complete Attribution

#### 5.3.1 Expected Inactivity Does Not Establish Equivalent Execution

The governance comparison asks whether a constraint that appears irrelevant can be treated as equivalent without running the constrained policy independently. Independent execution returned different model paths in 19 of 599 cases expected to remain unchanged, showing that theoretical nonbinding status cannot establish implementation equivalence. These path differences are detectable internal disagreements; the results do not imply that every disagreement caused meaningful plan or outcome divergence. In approximate search, the feasibility rule can change which partial paths remain available even when the final unconstrained solution satisfies the limit. Table 5 reports enforcement and cost trade-offs when the unconstrained path exceeds the limit.

Table 5. Expected constraint inactivity must be verified through independent execution.

| Dataset | Cap rule | K | Expected-binding rate | Switch reduction | Cost change | Interpretation |
|---|---|---|---|---|---|---|
| M5 | Fixed | 2 | 63% | 2.93 | −0.11% | Expected binding; negligible mean cost change. |
| M5 | Validation quartile | 2 | 63% | 2.93 | −0.11% | Validation rule selects the same cap. |
| Walmart | Fixed | 2 | 99% | 5.64 | +2.66% | 99% expected binding with measurable cost. |
| Walmart | Validation quartile | 6 | 65% | 2.03 | −0.07% | Looser cap removes the mean cost trade-off. |

*Note:* Expected binding: the independently computed unconstrained path exceeds $K$; switch and cost effects compare independent runs.

Expected inactivity can prioritize review, but only independent execution establishes equivalence.

#### 5.3.2 Proxy Agreement Can Hide Decision Risk

The diagnostic comparison asks whether a lower-cost proxy can safely replace full execution evaluation. Winner labels agreed in 55.0% of units, while the static choice was cost-equivalent within the $10^{-6}$ study tolerance in 79.5%. Thus, label disagreement and meaningful operational divergence are not equivalent concepts. Among the remaining non-equivalent choices, dataset-specific P90 raw regret was 14.1 for M5 and 41,569 for Walmart; this upper tail prevents the proxy from authorizing acceptance without independent confirmation. Figure 3 separates winner agreement, cost equivalence, and upper-tail regret.

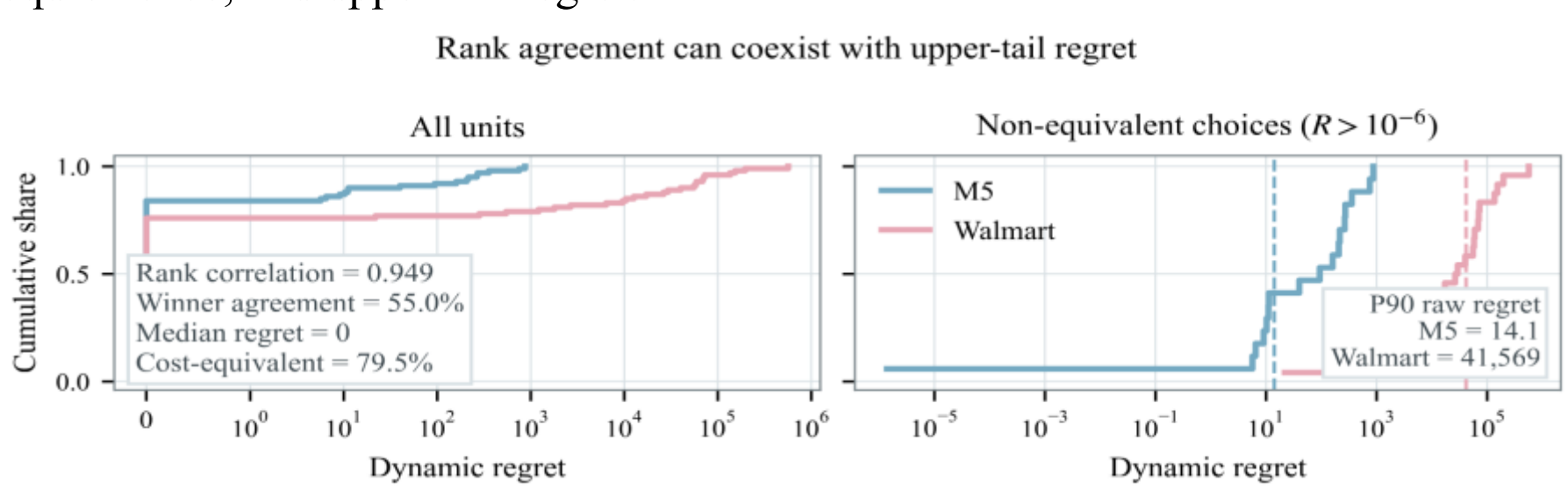

Figure 3. Frequent cost equivalence coexists with dataset-specific upper-tail regret.

The evidence-sufficiency objective therefore yields a bounded conclusion: indirect checks can prioritize review, but cannot replace independent confirmation when equivalence or adverse outcomes remain unresolved. Together with Sections 5.1 and 5.2, this establishes that *no single indicator considered in isolation is sufficient for capability-level operational attribution*. Aggregate improvement, executed-plan change, and indirect evidence address different parts of attribution and must remain jointly visible within an explicit operating scope.

### 5.4 Joint Evidence Determines Lifecycle Action

Figure 1 defines the evidence required before operational credit can be considered. Figure 4 shows how the joint pattern constrains organizational action and how monitoring updates the same record.

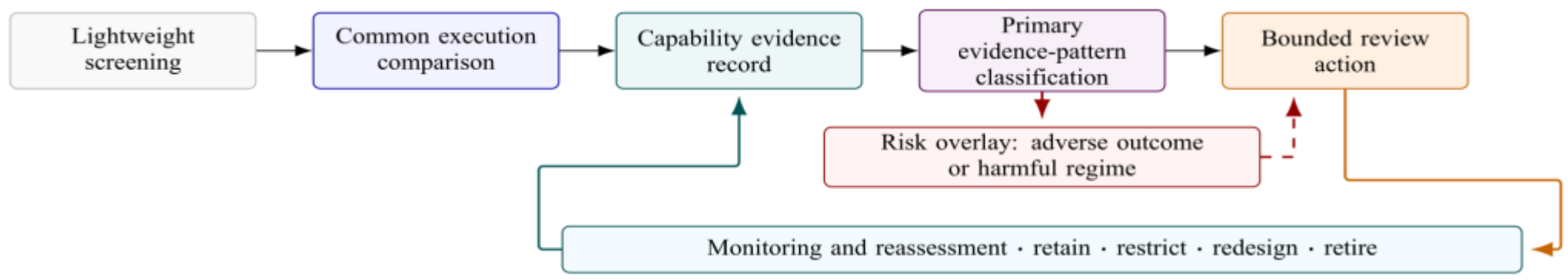

Figure 4. Joint evidence determines whether a capability advances, remains restricted, or requires further confirmation.

The capability evidence record is the standard's practical output, not a substitute for evaluation. It combines the frozen contract with observed opportunity, plan change, consequence, uncertainty, supported scope, adverse evidence, and unresolved findings. A *risk overlay* is a separate designation for an adverse outcome or harmful regime that meets the frozen adverse-risk criterion; it can modify any primary status. Organization-supplied burden completes the value decision. For Industrial AI deployment, this record makes user acceptance testing (UAT) eligibility and scope traceable to evidence. Figure 4 shows the process, whereas Table 6 provides the decision lookup.

Table 6. Different evidence patterns justify different lifecycle responses.

| **Evidence pattern** | **Status** | **Default lifecycle response** |
|---|---|---|
| No meaningful opportunity to act or bind | Inactive | Revisit the use case; assign no incremental credit |
| Plan and relevant outcomes unchanged | Operationally equivalent | Retain the simpler baseline unless another objective justifies the added complexity |
| Stable benefit within the declared scope | Supported | Proceed to monitored UAT or phased deployment |
| Benefit limited to identifiable regimes | Conditional | Restrict eligibility and retain baseline fallback |
| Evidence insufficient, conflicting, or imprecise | Unresolved | Collect execution evidence, redesign, or defer the decision |
| Adverse outcome or harmful regime meeting the frozen risk criterion | **Risk overlay** | Narrow scope and add guardrails, override, monitoring, and rollback |

*Note:* The adverse-risk criterion is frozen before evaluation for the capability's intended and protected outcomes; no universal severity threshold is implied. The overlay may modify any primary status.

### 5.4.1 Completed Evidence Record: Restrict, Test, and Reassess

**Contract and evidence.** The evaluated addition is the approximate multi-period capability, compared with the simple ensemble under frozen long-lead/carryover and tight-capacity scopes. Full-horizon replay provides favorable evidence in identified long-lead regimes and a 7.6% adverse result under M5 tight capacity. The tractable exact benchmark separately confirms execution fidelity of the approximate implementation within its reduced scope. **Interpretation and action.** The record assigns *Conditional* status with a *Risk* overlay. UAT is limited to eligible long-lead units, while the baseline remains active under tight capacity and outside the supported scope. Benefit decay, adverse drift, or burden disproportionate to contribution triggers restriction, redesign, or retirement.

## 5.5 What the Study Establishes—and What Remains Organizational

Within the sampled panels and declared scopes, replay identifies whether an addition influenced execution and whether its consequences support credit or leave equivalent, adverse, or uncertain evidence.

UAT must test whether contribution survives use; enterprise-calibrated value and sustaining burden require local evidence. Exactness is shown only for the reduced benchmark, and the within-window forecast-horizon counterfactual remains unresolved. The cases establish neither causal effects, enterprise prevalence, nor a universal deployment rule. Future work should validate the record in live workflows and develop leakage-safe screens under the same information boundary.

# 6 Conclusion

System-level evidence can misstate capability-level value: one control changed only 15 of 681 executed plans, and the evaluated multi-period capability became 7.6% more costly under tight capacity. Rule-based expectations and proxy agreement likewise cannot establish implementation equivalence in isolation.

Capability attribution is a sequence of evidence, not a single score. Internal-path agreement, executed-plan equivalence, and operational consequence must remain distinct within a declared scope. The evidence standard keeps these observations visible at the added-capability level, and its capability evidence record makes attribution and action traceable.

The record changes the organizational workflow. Inactive or equivalent additions receive no incremental credit. Conditional additions retain a baseline fallback; supported additions proceed through monitored UAT. Unresolved or adverse evidence requires confirmation, restriction, redesign, or retirement. Capabilities are added by design; credit must be earned. Operational value exists only when independent contribution is observable, defensible within a declared scope, and commensurate with the burden of continued use.

**Data and Code Availability.** Code, experiment configurations, canonical result tables, audit artifacts, and reproduction instructions are available at https://github.com/yvonnexu442/planning-capability-evidence-standard/releases/tag/v1.0-camera-ready. The Favorita, M5, and Walmart datasets are available from their cited original sources and are not redistributed.

## References

Ansari, A. F., Stella, L., Türkmen, C., Zhang, X., Mercado, P., Shen, H., Shchur, O., Rangapuram, S. S., Pineda Arango, S., Kapoor, S., Zschiegner, J., Maddix, D. C., Wang, H., Mahoney, M. W., Torkkola, K., Wilson, A. G., Bohlke-Schneider, M. and Wang, Y., Chronos: Learning the language of time series, *Transactions on Machine Learning Research*, October 2024.

Axsäter, S., *Inventory Control*, 3rd Edition, Springer, 2015.

Bates, J. M. and Granger, C. W. J., The combination of forecasts, *Operational Research Quarterly*, vol. 20, no. 4, pp. 451-468, 1969.

Culot, G., Podrecca, M. and Nassimbeni, G., Artificial intelligence in supply chain management: A systematic literature review of empirical studies and research directions, *Computers in Industry*, vol. 162, article 104132, 2024.

Das, A., Kong, W., Sen, R. and Zhou, Y., A decoder-only foundation model for time-series forecasting, *Proceedings of the 41st International Conference on Machine Learning*, PMLR, vol. 235, pp. 10148-10167, 2024.

Delage, E., Arroyo, S. and Ye, Y., The value of stochastic modeling in two-stage stochastic programs with cost uncertainty, *Operations Research*, vol. 62, no. 6, pp. 1377-1393, 2014.

Elmachtoub, A. N. and Grigas, P., Smart "Predict, then Optimize," *Management Science*, vol. 68, no. 1, pp. 9-26, 2022.

Forrester, J. W., *Industrial Dynamics*, MIT Press, Cambridge, MA, 1961.

Hyndman, R. J. and Koehler, A. B., Another look at measures of forecast accuracy, *International Journal of Forecasting*, vol. 22, no. 4, pp. 679-688, 2006.

Januschowski, T., Wang, Y., Torkkola, K., Erkkilä, T., Hasson, H. and Gasthaus, J., Forecasting with trees, *International Journal of Forecasting*, vol. 38, no. 4, pp. 1473-1481, 2022.

Jia, M., Schrotenboer, A. H. and Chen, F., Scenario predict-then-optimize for data-driven online inventory routing, *Transportation Science*, vol. 59, no. 5, pp. 1032-1056, 2025.

Li, X., Hu, Y., Su, X. and Shao, B., A data-driven newsvendor problem with shifting demand: A deep autoregressive model with attention mechanism, *Journal of Engineering Science and Technology Review*, vol. 16, no. 3, pp. 74-83, 2023.

Makridakis, S., Spiliotis, E. and Assimakopoulos, V., M5 accuracy competition: Results, findings, and conclusions, *International Journal of Forecasting*, vol. 38, no. 4, pp. 1346-1364, 2022.

Mandi, J., Kotary, J., Berden, S., Mulamba, M., Bucarey, V., Guns, T. and Fioretto, F., Decision-focused learning: Foundations, state of the art, benchmark and future opportunities, *Journal of Artificial Intelligence Research*, vol. 80, pp. 1623-1701, 2024.

Rice, J. R., The algorithm selection problem, *Advances in Computers*, vol. 15, pp. 65-118, 1976.

Sethi, S. P. and Sorger, G., A theory of rolling horizon decision making, *Annals of Operations Research*, vol. 29, no. 1, pp. 387-415, 1991.

Silver, E. A., Pyke, D. F. and Peterson, R., *Inventory Management and Production Planning and Scheduling*, 3rd Edition, John Wiley and Sons, New York, 1998.

Stadtler, H., Kilger, C. and Meyr, H., eds., *Supply Chain Management and Advanced Planning: Concepts, Models, Software, and Case Studies*, 5th Edition, Springer, 2015.

Su, X., Zeng, L., Shao, B. and Lin, B., Data-driven optimization for production planning with multiple demand features, *Kybernetes*, vol. 54, no. 1, pp. 110-133, 2025.

Vlachos, I. and Reddy, P. G., Machine learning in supply chain management: Systematic literature review and future research agenda, *International Journal of Production Research*, vol. 63, no. 16, pp. 5987-6016, 2025.

Wang, Y., Han, X., Shang, D., Tang, Y. and Liu, B., Safety, or just capability? A validity audit of agent-safety benchmarks, *arXiv preprint arXiv:2607.28685*, 2026.

Walter, A., Ahsan, K. and Rahman, S., Application of artificial intelligence in demand planning for supply chains: A systematic literature review, *The International Journal of Logistics Management*, vol. 36, no. 3, pp. 672-719, 2025.